\documentclass[a4paper,fleqn]{cas-sc}

\usepackage[numbers,sort&compress]{natbib}

\usepackage{enumitem}
\usepackage{mathtools}

\newcommand{\TwistNet}{TwistNet-2D\xspace}

\begin{document}

\let\WriteBookmarks\relax
\def\floatpagepagefraction{1}
\def\textpagefraction{.001}

\shorttitle{Spiral-Twisted Channel Interactions for Texture Recognition}
\shortauthors{J. J. Lian et~al.}

\title[mode=title]{TwistNet-2D: Learning Second-Order Channel Interactions via Spiral Twisting for Texture Recognition}

\author[1]{Junbo Jacob Lian}[%
        orcid=0000-0001-7602-0022]
\ead{junbolian@u.northwestern.edu}
\credit{Formal analysis, Conceptualization, Resources, Investigation, Methodology, Software, Funding acquisition, Writing - original draft, Writing - review \& editing.}

\author[1]{Feng Xiong}[orcid=0009-0003-8439-857X]
\ead{fengxiong2026@u.northwestern.edu}
\credit{Validation, Software, Investigation, Data curation, Resources}

\author[1]{Yujun Sun}[orcid=0009-0008-7336-8320]
\ead{yujunsun2026@u.northwestern.edu}
\credit{Investigation, Validation, Writing - review \& editing}

\author[2]{Kaichen Ouyang}[orcid=0009-0003-5937-5229]
\ead{oykc@mail.ustc.edu.cn}
\credit{Investigation, Resources, Writing - review \& editing}

\author[3]{Zong Ke}[orcid=0009-0006-0450-5614]
\ead{a0129009@u.nus.edu}
\credit{Investigation, Resources}

\author[4]{Mingyang Yu}[orcid=0009-0000-5903-6014]
\ead{1120240312@mail.nankai.edu.cn}
\credit{Investigation}

\author[5]{Shengwei Fu}[orcid=0000-0001-7424-0291]
\ead{gs.swfu22@gzu.edu.cn}
\credit{Investigation}

\author[6]{Rui Zhong}[orcid=0000-0003-4605-5579]
\ead{zhongrui@iic.hokudai.ac.jp}
\credit{Investigation}

\author[7]{Yujun Zhang}[orcid=0000-0003-3016-8843]
\ead{zhangyj069@gmail.com}
\credit{Investigation}

\author[8]{Huiling Chen}[orcid=0000-0002-7714-9693]
\cormark[1]
\ead{chenhuiling.jlu@gmail.com}
\credit{Supervision, Writing -- Review \& Editing}

\affiliation[1]{organization={McCormick School of Engineering, Northwestern University},
    city={Evanston},
    state={IL},
    country={USA}}

\affiliation[2]{organization={School of Mathematics, University of Science and Technology of China}, city={Hefei}, country={China}}

\affiliation[3]{organization={National University of Singapore},
    city={Singapore},
    country={Singapore}}

\affiliation[4]{organization={College of Artificial Intelligence, Nankai University},
               city={Tianjin},
               country={China}}

\affiliation[5]{organization={Guizhou University},
               city={Guiyang},
               country={China}}

\affiliation[6]{organization={Information Initiative Center, Hokkaido University}, city={Sapporo}, country={Japan}}

\affiliation[7]{organization={School of New Energy, Jingchu University of Technology}, city={Jingmen}, country={China}}

\affiliation[8]{organization={Wenzhou University},
    city={Wenzhou},
    country={China}}

\cortext[cor1]{Corresponding author.}

\begin{abstract}
Second-order feature statistics are central to texture recognition, yet
existing mechanisms exhibit a structural tension: bilinear pooling and Gram
matrices capture global channel correlations but discard spatial structure,
whereas self-attention models cross-position relations through weighted sums
rather than explicit pairwise products. We propose \TwistNet, a lightweight
module that computes local pairwise channel products under directional
spatial displacement, jointly encoding where features co-occur and how they
interact. The core component, Spiral-Twisted Channel Interaction (STCI),
shifts one feature map along a prescribed direction before
$\ell_2$-normalized channel multiplication, capturing the cross-position
co-occurrence patterns that characterize structured and periodic textures.
Four directional heads are aggregated through content-adaptive channel
reweighting, and the result is injected via a sigmoid-gated residual path
with near-zero initialization. \TwistNet adds only ${\sim}3.5\%$ parameters
and ${\sim}2\%$ FLOPs over ResNet-18. To isolate the contribution of
architectural inductive bias from that of transfer learning, all models in
this study are trained from scratch without ImageNet pretraining. Under this
protocol, \TwistNet consistently surpasses parameter-matched baselines and
substantially larger ConvNeXt and Swin Transformer backbones across four
texture and fine-grained recognition benchmarks, while the multi-head
structure produces interpretable, orientation-selective representations that
align with classical texture analysis. The open code will be available at \url{https://github.com/junbolian/TwistNet-2D}.
\end{abstract}

\begin{highlights}
\item Spiral-twisted directional pairwise products model cross-position co-occurrence.
\item MH-STCI module adds only $3.5\%$ parameters and $2\%$ FLOPs to ResNet-18.
\item \TwistNet-18 outperforms parameter-matched and $2.4\times$ larger baselines.
\item Ablations attribute the gain to local second-order modeling with AIS.
\item Directional heads acquire interpretable, orientation-selective behavior.
\end{highlights}

\begin{keywords}
Texture recognition \sep Second-order features \sep Channel interaction \sep
Fine-grained classification \sep Lightweight networks
\end{keywords}

\maketitle

\section{Introduction}
\label{sec:intro}

Texture recognition demands understanding \emph{how} visual features
relate, not merely \emph{what} features are present. Wood grain illustrates
the point: its identity is defined not by the presence of dark stripes or
light bands in isolation but by their alternation at regular spatial
offsets. The classical formulation of this insight is Haralick's
gray-level co-occurrence matrix~\cite{haralick1973textural}, which counts
feature pairs at specified spatial displacements; the modern challenge is
to incorporate the same principle---explicit, spatially-aware
co-occurrence modeling---into deep representations without sacrificing
the flexibility of learned features.

Existing deep approaches to second-order modeling each capture some part
of this principle while discarding another. Bilinear
pooling~\cite{lin2015bilinear,gao2016compact} and Gram
matrices~\cite{gatys2016image} compute products
$\sum_{x,y} z_i(x,y)\, z_j(x,y)$ summed over all positions, retaining
\emph{which} channels co-occur but losing \emph{where} they do so. Their
local counterparts compute the same product per position, preserving
spatial structure but limiting interaction to coincident features---an
artifact that misaligns with periodic textures, in which correlated
features systematically appear at \emph{neighboring} positions. Self-
attention~\cite{vaswani2017attention} models cross-position relationships
but through weighted sums $\sum_j \alpha_{ij} v_j$, which recombine
existing features rather than form new second-order ones. The structural
gap is therefore specific: a local operator that takes pairwise
\emph{products} between features at \emph{displaced} positions, and
thereby unifies the per-position spatial preservation of local bilinear
pooling with the cross-position structure of co-occurrence analysis.

We propose \textbf{Spiral-Twisted Channel Interaction (STCI)}, an operator
that fills exactly this gap. STCI computes
\begin{equation}
\phi_{ij}(x,y) \;=\; \bar{z}_i(x,y) \cdot \bar{z}_j(x{+}\delta_x, y{+}\delta_y),
\label{eq:phi}
\end{equation}
where $\bar{z}$ denotes the channel-wise $\ell_2$-normalized feature and
the displacement $(\delta_x, \delta_y)$ is realized by a $3{\times}3$
depthwise convolution whose kernel is initialized to sample from a chosen
direction. Four directions
($0^\circ, 45^\circ, 90^\circ, 135^\circ$) span the orientation range of
plane textures, and the multi-directional outputs are aggregated through
an \textbf{Adaptive Interaction Selection (AIS)} module that places
SE-style attention~\cite{hu2018squeeze} \emph{on the resulting interaction
channels} rather than on raw first-order features, allowing the network
to route each image through whichever co-occurrence pattern is most
informative. A sigmoid-gated residual connection initialized near zero
inserts the resulting MH-STCI branch into a ResNet block without
disrupting early-stage optimization. The composite operator is
parameter-efficient: \TwistNet-18 adds ${\sim}3.5\%$ parameters and
${\sim}2\%$ FLOPs over ResNet-18.

The construction admits an information-theoretic interpretation. For
jointly Gaussian features, mutual information depends only on the
correlation coefficient,
$I(Z_i; Z_j) = -\tfrac{1}{2}\log(1{-}\rho_{ij}^2)$, and the
$\ell_2$-normalized products in Eq.~\ref{eq:phi} are pointwise estimators
of $\rho_{ij}$ at displaced positions. Explicit pairwise products
therefore compute the statistical quantity that the texture-recognition
problem requires, rather than relying on the network to discover it
through gradient descent across a much larger hypothesis class.

All experiments in this paper use a single, deliberately controlled
protocol: every model---ours and every baseline---is trained from scratch
on the target dataset, with no ImageNet pretraining. Pretrained
backbones combined with texture-specific encoders such as
DEP~\cite{xue2018deep}, DeepTEN~\cite{zhang2017deepten},
MAP-Net~\cite{zhai2020dsrnet,zhai2023multiplicity}, and their recent
extensions~\cite{evani2025chebyshev,sikdar2025interweaving} routinely
exceed $75\%$ on DTD and $90\%$ on CUB-200, but in those pipelines the
texture-specific module's contribution is inseparable from the
representation transferred from large-scale pretraining. The from-scratch
protocol isolates what an operator can express from gradient signal on a
small target dataset; the absolute accuracies are correspondingly lower
than the pretrained literature, and we contextualize the gap
quantitatively in Section~\ref{sec:protocol}.

\paragraph{Contributions.}
\begin{itemize}[leftmargin=*,itemsep=0.15em]
\item We propose Spiral-Twisted Channel Interaction, a directionally-
displaced second-order operator that extends per-position pairwise
products into cross-position co-occurrence detectors and supplies the
local Haralick-style structure missing from existing global-bilinear and
self-attentive designs (Section~\ref{sec:stci}).
\item We design a multi-head architecture with Adaptive Interaction
Selection and gated residual injection that aggregates four directional
heads into an integrated MH-STCI module, applies channel attention to
second-order interaction features rather than first-order ones, and
inserts cleanly into a standard ResNet without disrupting backbone
optimization (Sections~\ref{sec:mhstci} and \ref{sec:twistblock}).
\item Under a uniform from-scratch protocol on four texture and
fine-grained recognition benchmarks, \TwistNet-18 consistently outperforms
parameter-matched baselines and substantially larger ConvNeXt and Swin
Transformer backbones at $3.5\%$ parameter and $2\%$ FLOP overhead over
ResNet-18; a component-level decomposition relates the operator to the
local-bilinear baseline it generalizes, and a regime analysis explains
where cross-position structure carries class identity
(Section~\ref{sec:experiments}).
\end{itemize}

\section{Related Work}
\label{sec:related}

\paragraph{Second-order representations.}
The role of second-order statistics in texture is established
theoretically~\cite{julesz1981textons,portilla2000parametric} and
practically~\cite{haralick1973textural}. Gram matrices match global
texture statistics for style transfer~\cite{gatys2016image,
johnson2016perceptual,li2017demystifying}; bilinear pooling computes outer
products for fine-grained recognition~\cite{lin2015bilinear}, with efficient
variants via compact projections~\cite{gao2016compact}, low-rank
factorization~\cite{kong2017lowrank}, and matrix-power
normalization~\cite{li2017second}; recent extensions explore higher-order
interactions~\cite{sikdar2025interweaving}. The structural feature shared
by these methods is that pooling occurs before or simultaneously with the
second-order product: information is first aggregated globally, and only
then are pairwise statistics formed. This is appropriate for
translation-invariant style matching but discards the spatial periodicity
that characterizes Brodatz-style and natural textures. The local,
cross-position products studied here provide complementary information:
not only that features co-occur, but where on the image and at what
spatial offset.

\paragraph{Texture recognition.}
Deep texture recognition has progressed from CNN features with orderless
pooling~\cite{cimpoi2014dtd,cimpoi2015deep} to specialized architectures.
DEP~\cite{xue2018deep} learns a deep encoding pooling layer over
pretrained convolutional features for ground-terrain recognition;
DeepTEN~\cite{zhang2017deepten} introduces a learnable residual encoding
that integrates texton-style aggregation with deep features. The
DSR-Net family~\cite{zhai2020dsrnet} aggregates multi-scale structural
information, and~\cite{zhai2023multiplicity} extends it through a notion
of primitive multiplicity for in-the-wild textures. More recently, an
interweaving of bilinear and tri-linear
interactions~\cite{sikdar2025interweaving} captures higher-order channel
relations in fine-grained recognition, while spectral attention via
Chebyshev polynomial bases~\cite{evani2025chebyshev} provides a spectral
counterpart to spatial-domain attention. Recent contributions in
\emph{Pattern Recognition} pursue complementary aggregation strategies
on top of pretrained backbones: RADAM~\cite{scabini2023radam} encodes
activation maps from a frozen CNN at multiple depths through a
randomised autoencoder, producing a single texture descriptor without
fine-tuning; deep tracing pattern encoding~\cite{chen2024tracing} traces
the features generated along convolutional layers via histograms of
local 3D rotation-invariant binary patterns to construct a cross-layer
global representation; and \cite{shu2022global} refines local binary
descriptors with global neighbourhood statistics for texture
classification. The shared structural choice in all of these methods is
to aggregate features---through pooling, encoding, attention, or
histogramming---on top of a pretrained or otherwise fixed backbone, so
that the downstream module enriches representations whose spatial
geometry is already imprinted by ImageNet-scale training or by
handcrafted local descriptors. Our route is orthogonal: we modify the
convolutional block itself with an in-backbone second-order operator,
so that cross-position co-occurrence is modelled \emph{during}
representation learning rather than aggregated \emph{after} it. Because
prior contributions are reported under pretrained or descriptor-coupled
protocols, their absolute numbers are not directly comparable with the
from-scratch numbers reported in Section~\ref{sec:experiments}; we
return to this point in Section~\ref{sec:protocol}.

\paragraph{Attention mechanisms.}
Self-attention~\cite{vaswani2017attention} was adapted to vision via
ViT~\cite{dosovitskiy2021vit} and refined with local
windows~\cite{liu2021swin}, linear approximations~\cite{katharopoulos2020transformers},
and hybrid CNN--transformer designs~\cite{vasu2023fastvit,tu2022maxvit}.
Channel attention~\cite{hu2018squeeze,woo2018cbam,wang2020eca,
hou2021coordinate} reweights features through global statistics. The
distinction with the present work is operator-level: attention computes
weighted sums over existing features, whereas STCI computes products that
create new interaction features. The AIS module then applies channel
attention on top of these interaction features. The novelty of AIS
therefore does not lie in the squeeze--excite mechanism itself but in the
object on which it operates, namely a multi-directional second-order
interaction field.

\paragraph{Polynomial and twisted networks.}
Modern CNNs~\cite{he2016deep,liu2022convnet,tan2019efficientnet,
wang2024repvit} provide strong baselines, and STCI is designed as a
complementary module insertable with minimal modification. Polynomial
networks~\cite{chrysos2021deep} motivate explicit higher-order expansions,
and STCI may be viewed as a spatially-aware quadratic expansion with
aggressive bottlenecking. Twisted interactions were previously introduced
for tabular data~\cite{lian_tcn}; the present work generalizes the
operator to the 2D image domain, where the ``twist'' is redefined as
directional spatial displacement targeting the cross-position structures
that characterize visual textures.

\section{Method}
\label{sec:method}

\begin{figure*}[t]
\centering
\includegraphics[width=0.95\textwidth]{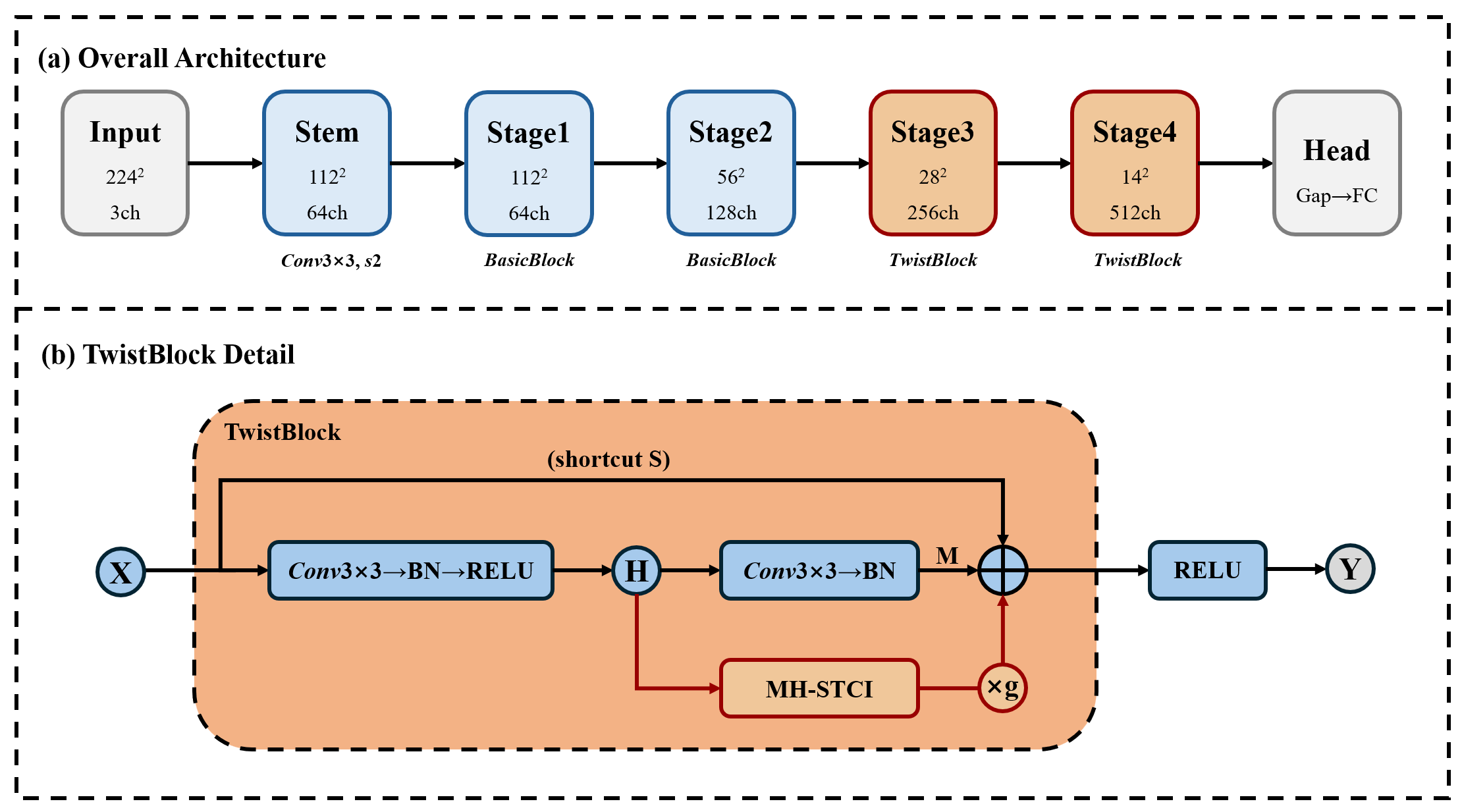}
\caption{\textbf{\TwistNet architecture.} Top: \TwistNet-18 follows a
ResNet-like structure with four stages; Stages~3--4 use TwistBlocks that
inject second-order channel interactions through a gated MH-STCI branch.
Bottom: a TwistBlock augments the standard residual block with the gated
MH-STCI branch operating on the intermediate activation $H$. The internal
structure of MH-STCI and a single STCI head is detailed in
Figs.~\ref{fig:mhstci} and \ref{fig:stci}.}
\label{fig:overview}
\end{figure*}

Figure~\ref{fig:overview} illustrates \TwistNet at three levels. We first
present the theoretical motivation (Section~\ref{sec:motivation}), then
describe the STCI head (Section~\ref{sec:stci}), the multi-head aggregation
with adaptive selection (Section~\ref{sec:mhstci}), the residual integration
(Section~\ref{sec:twistblock}), and finally the full architecture
(Section~\ref{sec:arch}).

Throughout, $X \in \mathbb{R}^{C \times H \times W}$ denotes an intermediate
feature map. We use $z_i \in \mathbb{R}^{H \times W}$ for the $i$-th channel,
$\bar{z}_i$ for its $\ell_2$-normalized version, and $\tilde{z}_i$ for its
directionally displaced version. The four canonical directions are indexed
by $\theta \in \{0, \pi/4, \pi/2, 3\pi/4\}$. Symbols and shapes are
summarized in Table~\ref{tab:notation}.

\begin{table}[t]
\centering
\caption{Symbols used in the description of the method. $C_r$ is the
reduced channel count (default 8); $K=4$ is the number of directional
heads; $P=C_r(C_r{+}1)/2$ is the number of upper-triangular pairwise
products.}
\label{tab:notation}
\small
\setlength{\tabcolsep}{4pt}
\begin{tabular}{l l l}
\toprule
\textbf{Symbol} & \textbf{Shape} & \textbf{Meaning} \\
\midrule
$X$ & $C\!\times\! H\!\times\! W$ & Input feature map to a TwistBlock \\
$Z$ & $C_r\!\times\! H\!\times\! W$ & Channel-reduced features \\
$\tilde{Z}_\theta$ & $C_r\!\times\! H\!\times\! W$ & Directionally shifted features \\
$\bar{Z}, \bar{\tilde{Z}}_\theta$ & $C_r\!\times\! H\!\times\! W$ & $\ell_2$-normalized features \\
$\phi_\theta$ & $P\!\times\! H\!\times\! W$ & Pairwise product field for direction $\theta$ \\
$F$ & $D\!\times\! H\!\times\! W$ & Concatenated $K$ heads, $D = K(C_r{+}P) = 176$ \\
$g = \sigma(\gamma)$ & scalar & Gate, initialized at $\gamma{=}{-}2$ ($g\!\approx\! 0.12$) \\
\bottomrule
\end{tabular}
\end{table}

\subsection{Motivation}
\label{sec:motivation}

\begin{figure}[t]
\centering
\includegraphics[width=\columnwidth]{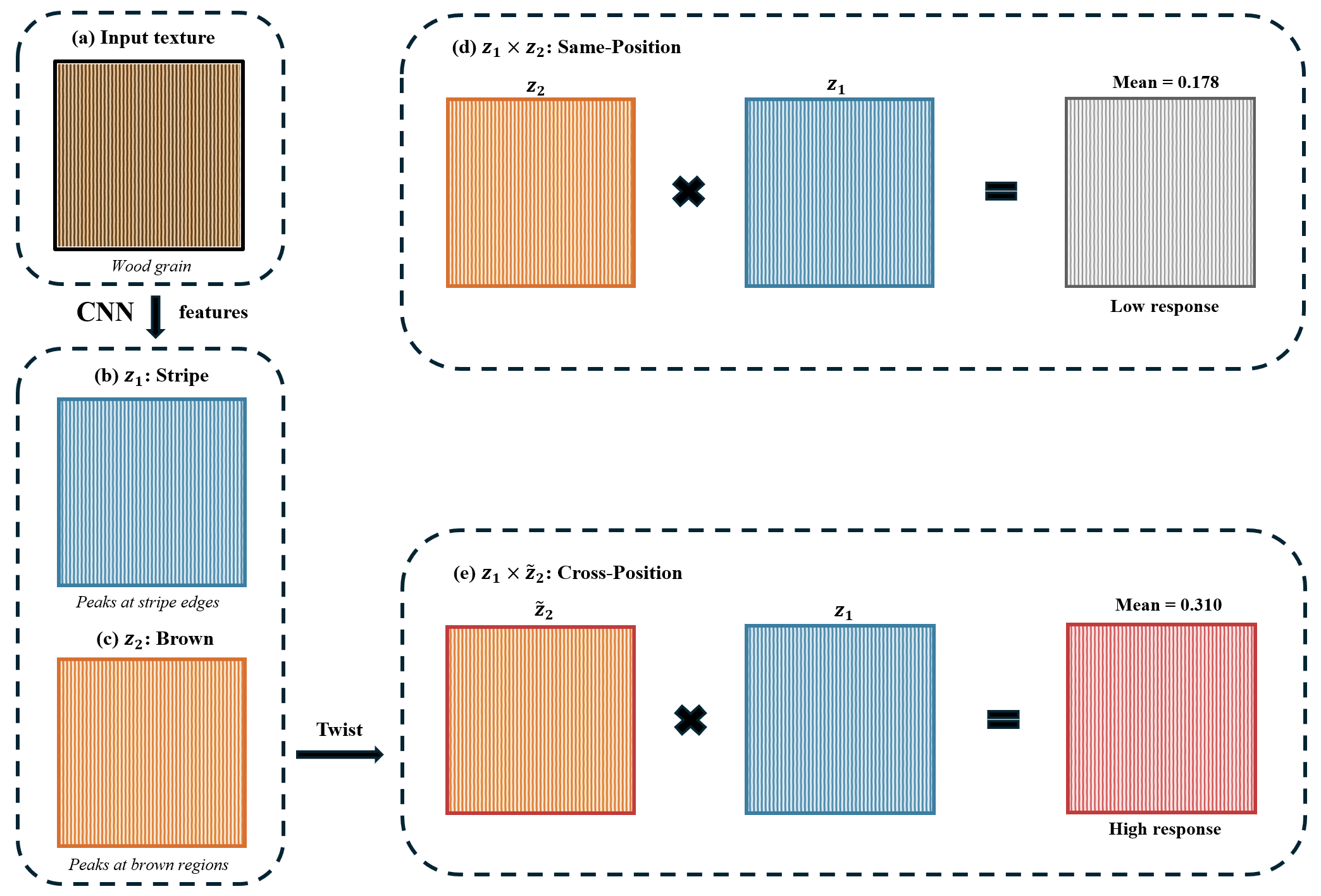}
\caption{\textbf{Why cross-position correlation?} (a) Wood grain exhibits
periodic stripe-brown alternation. (b)--(c) The CNN extracts a stripe
detector $z_1$ and a brown-region detector $z_2$ that respond at
interleaved positions. (d) The same-position product $z_1 z_2$ has a low
response because peaks of $z_1$ and $z_2$ are misaligned. (e) Spiral
Twist shifts $z_2$ by a small offset $\delta$ before the product; this
re-aligns peaks and recovers the periodic correlation.}
\label{fig:cross_position}
\end{figure}

The introduction sketched the structural gap that STCI addresses; here we
make the geometry of the operator concrete. The classical Gram matrix
$G_{ij} = \sum_{hw} F_{ihw} F_{jhw}$~\cite{gatys2016image} captures the
co-occurrence statistics of channels but pools across all spatial
positions, so the location and spatial relationship of co-occurring
features is discarded. Replacing the global sum with per-position
products $z_i(x,y) z_j(x,y)$ retains a spatial map of co-occurrence
intensity, but, restricted to coincident positions, it cannot represent
the spatial periodicity that distinguishes a stripe pattern from a
flecked one. The displaced product
$z_i(x,y) \tilde{z}_j(x{+}\delta_x, y{+}\delta_y)$ closes this gap.
Figure~\ref{fig:cross_position} shows the canonical case:
peaks of a stripe detector $z_1$ and of a brown-region detector $z_2$
alternate spatially, so their same-position product is small; a one-pixel
shift of $z_2$ aligns the peaks and recovers the periodic correlation
that defines wood grain. This is a learnable, end-to-end differentiable
analogue of Haralick's gray-level co-occurrence
matrix~\cite{haralick1973textural}, with the displacement playing the role
of the offset vector and the channel-wise normalization playing the role
of the histogram normalization.

\subsection{Spiral-Twisted Channel Interaction}
\label{sec:stci}

\begin{figure}[t]
\centering
\includegraphics[width=\columnwidth]{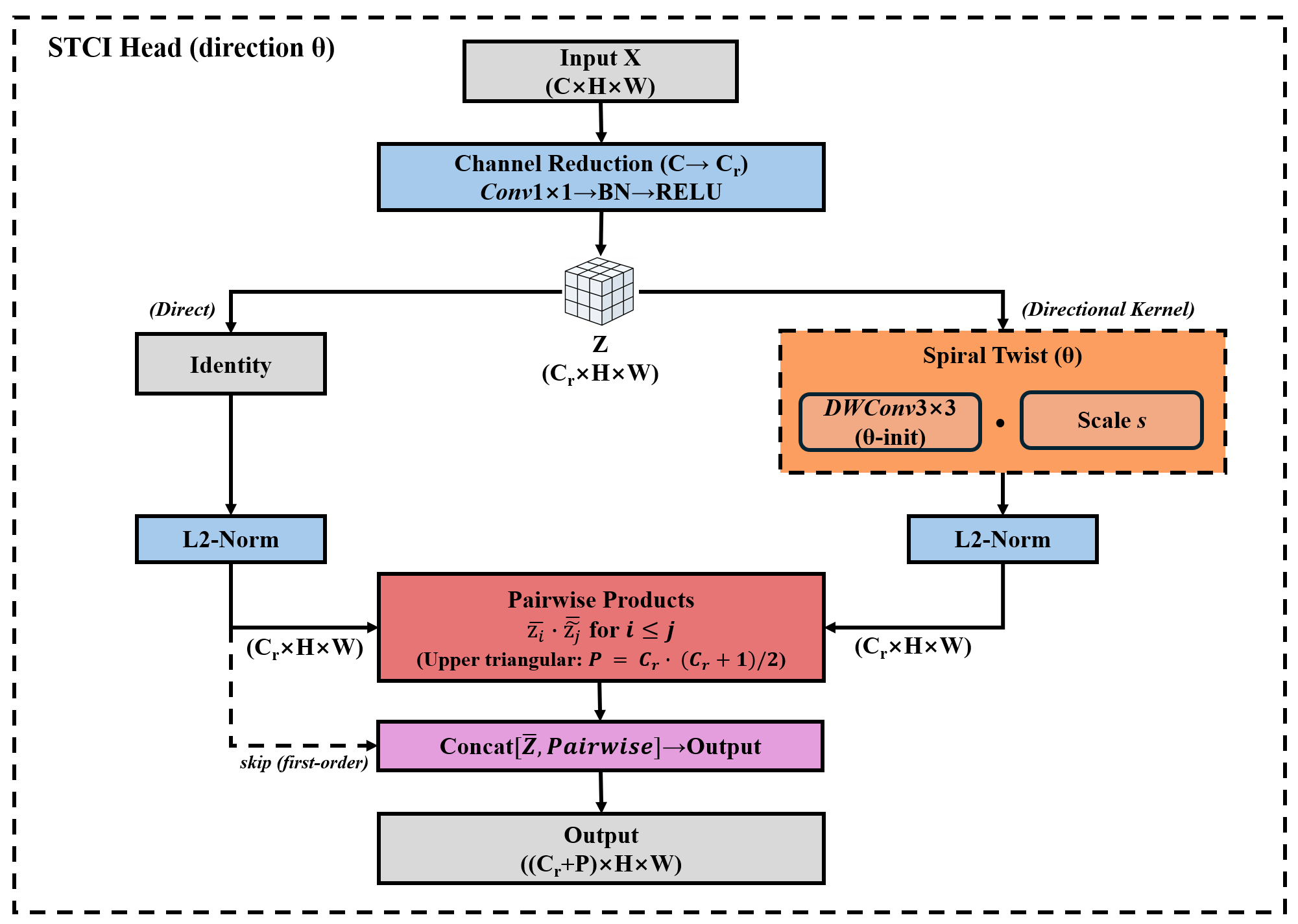}
\caption{\textbf{Single STCI head.} Channel reduction $C \to C_r$,
directional spiral twist via a fixed-pattern depthwise $3{\times}3$
convolution with a learnable per-channel scale, $\ell_2$-normalization
along the channel dimension, upper-triangular pairwise product field
$\phi_\theta \in \mathbb{R}^{P \times H \times W}$, and concatenation
with the normalized first-order features $\bar{Z}$.}
\label{fig:stci}
\end{figure}

A single STCI head, illustrated in Fig.~\ref{fig:stci}, computes local
second-order interactions in four steps.

\paragraph{Channel reduction.}
A $1{\times}1$ convolution compresses channels before the pairwise
expansion,
\begin{equation}
Z \;=\; \mathrm{ReLU}\bigl(\mathrm{BN}(W_{\mathrm{red}} * X)\bigr), \quad
W_{\mathrm{red}} \in \mathbb{R}^{C_r \times C \times 1 \times 1},
\label{eq:reduce}
\end{equation}
with $C_r \ll C$ (we use $C_r = 8$). The reduction lowers the cost of the
pairwise expansion from $O(C^2)$ to $O(C_r^2)$.

\paragraph{Directional spiral twist.}
A direction-specific depthwise convolution applies the spatial displacement,
\begin{equation}
\tilde{Z}_\theta \;=\; \mathrm{DWConv}_\theta(Z) \odot s,
\label{eq:twist}
\end{equation}
where $s \in \mathbb{R}^{C_r}$ is a learnable per-channel scale (initialized
to ones). Each kernel
$\mathbf{K}^\theta \in \mathbb{R}^{3 \times 3}$ is initialized to interpolate
between the center and one directionally-offset neighbor,
\begin{equation}
\mathbf{K}^\theta[1,1] = 0.5, \quad
\mathbf{K}^\theta[1{+}\delta_y, 1{+}\delta_x] = 0.5,
\label{eq:kernel_init}
\end{equation}
with all other entries zero, where
$(\delta_x, \delta_y) = (\mathrm{round}\cos\theta, \mathrm{round}\sin\theta)$
for $\theta \in \{0, \pi/4, \pi/2, 3\pi/4\}$. Table~\ref{tab:kernel} shows
the resulting kernels. The soft blend serves three purposes: it is a strict
generalization of the identity ($\delta_x = \delta_y = 0$ recovers the
same-position operator), it admits a non-zero gradient with respect to
position so the optimiser can refine the displacement, and the kernel
remains a learnable depthwise convolution rather than a fixed shift, so
heads are not constrained to remain at canonical angles after training.

\begin{table}[t]
\centering
\caption{Directional kernel initialization. Each $3{\times}3$ kernel
samples from the center (weight $0.5$) and one neighbor (weight $0.5$)
in the indicated direction.}
\label{tab:kernel}
\small
\setlength{\tabcolsep}{4pt}
\begin{tabular}{cccc}
\toprule
$\theta=0$ & $\theta=\pi/4$ & $\theta=\pi/2$ & $\theta=3\pi/4$ \\
\midrule
$\begin{bmatrix} 0 & 0 & 0 \\ 0 & .5 & .5 \\ 0 & 0 & 0 \end{bmatrix}$ &
$\begin{bmatrix} 0 & 0 & 0 \\ 0 & .5 & 0 \\ 0 & 0 & .5 \end{bmatrix}$ &
$\begin{bmatrix} 0 & 0 & 0 \\ 0 & .5 & 0 \\ 0 & .5 & 0 \end{bmatrix}$ &
$\begin{bmatrix} 0 & 0 & 0 \\ 0 & .5 & 0 \\ .5 & 0 & 0 \end{bmatrix}$ \\
($\rightarrow$) & ($\searrow$) & ($\downarrow$) & ($\swarrow$) \\
\bottomrule
\end{tabular}
\end{table}

We use four directions spanning $[0, \pi)$ rather than all eight compass
directions. After channel-wise $\ell_2$ normalization, the signed product
$\bar{z}_i \cdot \bar{\tilde{z}}_j$ measures correlation, and correlation
is symmetric under reflection: sampling along $\theta$ and along
$\theta + \pi$ yields equivalent statistics modulo a swap of indices
$i \leftrightarrow j$. The upper-triangular pairwise expansion (described
below) already enumerates both index orderings, so opposite directions are
redundant. Four directions therefore cover horizontal, vertical, and both
diagonal correlation axes while halving the cost of an eight-direction
scheme.

\paragraph{Normalization and pairwise products.}
Features are normalized channel-wise to stabilize magnitudes and ensure
that products estimate correlations,
\begin{equation}
\bar{Z} \;=\; \frac{Z}{\|Z\|_2 + \epsilon}, \qquad
\bar{\tilde{Z}}_\theta \;=\; \frac{\tilde{Z}_\theta}{\|\tilde{Z}_\theta\|_2 + \epsilon},
\label{eq:normalize}
\end{equation}
with the norm computed along the channel dimension at each spatial
location ($\epsilon = 10^{-6}$). The upper-triangular pairwise products are
then
\begin{equation}
\phi_\theta(\bar{Z}, \bar{\tilde{Z}}_\theta) \;=\;
\bigl[\bar{z}_i \cdot \bar{\tilde{z}}_j\bigr]_{1 \le i \le j \le C_r}
\;\in\; \mathbb{R}^{P \times H \times W},
\label{eq:pairwise}
\end{equation}
with $P = C_r(C_r{+}1)/2 = 36$ for $C_r = 8$. The diagonal terms
$\bar{z}_i \cdot \bar{\tilde{z}}_i$ measure auto-correlation across
displacement and are retained.

\paragraph{Head output.}
A single head outputs the concatenation of first-order and second-order
features,
\begin{equation}
\mathrm{STCI}_\theta(X) \;=\;
\bigl[\bar{Z} \,;\, \phi_\theta(\bar{Z}, \bar{\tilde{Z}}_\theta)\bigr]
\;\in\; \mathbb{R}^{(C_r + P) \times H \times W}.
\label{eq:stci_output}
\end{equation}
Including $\bar{Z}$ allows the module to fall back to linear behavior
when interactions are not beneficial. In practice the pairwise products
are implemented through precomputed upper-triangular index tensors, so
no explicit Python loop is required.

\subsection{Multi-Head Aggregation with Adaptive Selection}
\label{sec:mhstci}

\begin{figure}[t]
\centering
\includegraphics[width=\columnwidth]{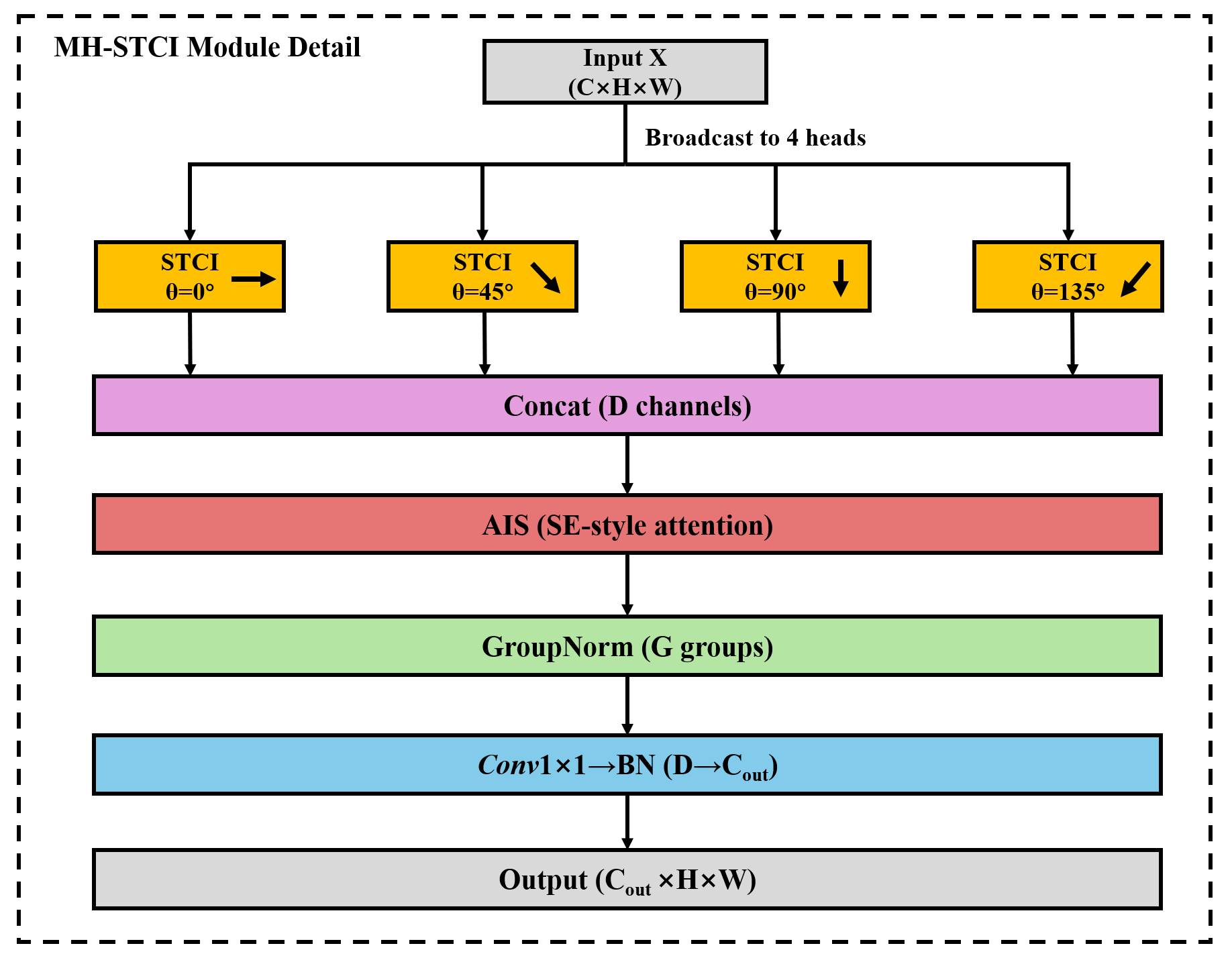}
\caption{\textbf{Multi-Head STCI.} Four directional heads
(Fig.~\ref{fig:stci}) are concatenated along the channel dimension,
reweighted by AIS with channel attention computed from the concatenated
interaction channels, normalized by GroupNorm, and projected to
$C_{\mathrm{out}}$ channels by a $1{\times}1$ convolution.}
\label{fig:mhstci}
\end{figure}

To achieve robustness across orientations, four directional STCI heads are
aggregated. With
$\theta_k \in \{0, \pi/4, \pi/2, 3\pi/4\}$,
\begin{equation}
F \;=\; \bigoplus_{k=1}^{K} \mathrm{STCI}_{\theta_k}(X)
\;\in\; \mathbb{R}^{D \times H \times W},
\label{eq:concat}
\end{equation}
where $D = K(C_r + P) = 4 \times 44 = 176$.

Adaptive Interaction Selection reweights the interaction channels through
an SE-style mechanism,
\begin{equation}
\mathrm{AIS}(F) \;=\; F \odot \sigma\bigl(W_2 \cdot \mathrm{ReLU}(W_1 \cdot \mathrm{GAP}(F))\bigr),
\label{eq:ais}
\end{equation}
with $W_1 \in \mathbb{R}^{D_{\mathrm{mid}} \times D}$,
$W_2 \in \mathbb{R}^{D \times D_{\mathrm{mid}}}$, and
$D_{\mathrm{mid}} = \max(D/4, 16)$. The structural shape of AIS deliberately
mirrors SE~\cite{hu2018squeeze}, but the input is categorically different:
SE operates on first-order features and reweights what is present, whereas
AIS operates on multi-directional second-order features and reweights
which co-occurrence pattern matters. The two mechanisms are complementary,
and the gap reported in Section~\ref{sec:main_results} between SE-ResNet-18
and \TwistNet under matched SE block geometry (Table~\ref{tab:main}) is
direct evidence that the operative variable is the object on which the
squeeze--excite acts, not the squeeze--excite mechanism itself.

GroupNorm~\cite{wu2018group} stabilizes optimization across batch sizes,
and a $1{\times}1$ projection produces the output:
\begin{equation}
\mathrm{MH\text{-}STCI}(X) \;=\; \mathrm{BN}\bigl(W_{\mathrm{proj}} *
\mathrm{GN}(\mathrm{AIS}(F))\bigr).
\label{eq:mhstci}
\end{equation}
The projection is initialized with small random values
($\mathrm{std} = 0.01$), so the output is approximately zero at
initialization; combined with the gate described next, this allows STCI
to be inserted into a backbone without disrupting initial training
dynamics.

\subsection{Gated Residual Integration}
\label{sec:twistblock}

The MH-STCI branch is integrated into a ResNet BasicBlock through a
sigmoid-gated injection. With
$H = \mathrm{ReLU}(\mathrm{BN}(W_1 * X))$ and
$M = \mathrm{BN}(W_2 * H)$ as the standard intermediate and main-path
activations,
\begin{equation}
Y \;=\; \mathrm{ReLU}\bigl(M + g \cdot \mathrm{MH\text{-}STCI}(H) + S(X)\bigr),
\label{eq:twistblock}
\end{equation}
where $S(\cdot)$ is identity or a $1{\times}1$ strided convolution and
$g = \sigma(\gamma)$ is a learned scalar. Initializing $\gamma = -2$ yields
$g \approx 0.12$ at the start of training, so the second-order contribution
is roughly an order of magnitude smaller than the main path and the block
behaves nearly identically to a standard ResNet block. During training the
gate is free to grow if interactions are beneficial; in our trained models
we observe that gates in Stage~3 stabilize around $0.25$--$0.40$ and gates
in Stage~4 around $0.35$--$0.55$ across runs, indicating that the network
voluntarily routes a non-trivial but bounded fraction of signal through
the second-order branch, with stronger reliance at deeper stages.

\subsection{Architecture and Complexity}
\label{sec:arch}

\TwistNet-18 follows the ResNet-18 layout: a standard stem ($7{\times}7$
convolution followed by $3{\times}3$ max-pool), four stages with channel
widths $(64, 128, 256, 512)$ and two blocks per stage, global average
pooling, and a linear classifier. Stages~1--2 use standard BasicBlocks;
Stages~3--4 use TwistBlocks. This placement reflects the texture-theoretic
observation that co-occurrence statistics are most informative among
mid-to-high-level units (textons, part-level fragments, repeated
microstructures)~\cite{julesz1981textons,portilla2000parametric} rather
than among raw edges. Inserting STCI earlier than Stage~3 produces
redundant correlations among low-level features; inserting only at Stage~4
misses the resolution at which structured periodicities such as fabric
weaves and feather barbs are most clearly captured.

Each TwistBlock adds approximately 70K parameters for its MH-STCI branch
(channel reduction, four spiral-twist depthwise convolutions, AIS,
projection). With four TwistBlocks in Stages~3--4, the total overhead is
${\sim}0.4$M parameters, bringing \TwistNet-18 to 11.59M parameters
(${\sim}3.5\%$ above ResNet-18's 11.20M). At Stage~3--4 resolutions
($28{\times}28$ and $14{\times}14$), the cost of the pairwise products is
small relative to the main-path $3{\times}3$ convolutions, and overall
FLOPs increase by ${\sim}2\%$ from 1.82G to 1.85G. The directional
depthwise convolutions and the upper-triangular pairwise products are
fully fused operations supported natively in PyTorch, and we observe
training-step wall-clock times within a few percent of unmodified
ResNet-18 on a single NVIDIA GPU, consistent with the reported FLOP
overhead.

\section{Experiments}
\label{sec:experiments}

We evaluate \TwistNet on four benchmarks under a uniform from-scratch
protocol, and we use the resulting numbers to address four questions:
whether explicit second-order modeling improves over first-order
baselines when transfer learning is removed, how \TwistNet compares to
recent efficient and high-capacity architectures under the same protocol,
which components of \TwistNet contribute to its performance, and whether
the directional heads acquire interpretable orientation selectivity.

\subsection{Experimental Setup}
\label{sec:setup}

\paragraph{Datasets.}
DTD~\cite{cimpoi2014dtd} (5{,}640 images, 47 texture classes, 10-fold CV);
FMD~\cite{sharan2009fmd} (1{,}000 images, 10 materials, 5-fold CV);
CUB-200~\cite{wah2011cub} (11{,}788 images, 200 bird species, 5-fold CV);
and Flowers-102~\cite{nilsback2008flowers} (8{,}189 images, 102 categories,
5-fold CV).

\paragraph{Baselines.}
The parameter-matched group (10--16M parameters) comprises
ResNet-18~\cite{he2016deep},
SE-ResNet-18~\cite{hu2018squeeze},
ConvNeXtV2-Nano~\cite{woo2023convnextv2},
FastViT-SA12~\cite{vasu2023fastvit}, and
RepViT-M1.5~\cite{wang2024repvit}; the larger-baseline group
(${\sim}28$M) comprises ConvNeXt-Tiny~\cite{liu2022convnet} and
Swin-Tiny~\cite{liu2021swin}. All baselines use the official
\texttt{timm}~\cite{wightman2021timm} implementations with only the
classification head re-initialized.

\paragraph{Training protocol.}
All models are trained from scratch on the target dataset, with no ImageNet
pretraining, under identical settings: 200 epochs, SGD with Nesterov
momentum ($0.9$), weight decay $10^{-4}$, cosine learning-rate schedule
(initial $0.05$, 10-epoch linear warmup), batch size 64, input
$224{\times}224$, mixed precision, and gradient clipping at $1.0$.
Augmentation includes RandomResizedCrop ($0.2$--$1.0$, BICUBIC),
RandomHorizontalFlip, RandAugment~\cite{cubuk2020randaugment}
($n{=}2, m{=}9$), Mixup~\cite{zhang2018mixup} ($\alpha{=}0.8$),
CutMix~\cite{yun2019cutmix} ($\alpha{=}1.0$), and label smoothing
($\epsilon{=}0.1$). Reported numbers are mean accuracy $\pm$ one standard
deviation across all folds and three random seeds (42, 43, 44).

\subsection{From-Scratch Protocol and Comparison with Pretrained Literature}
\label{sec:protocol}

The absolute accuracies in Table~\ref{tab:main} are well below those
reported elsewhere for the same benchmarks: pretrained pipelines combining
a ResNet-50 backbone with FV-CNN~\cite{cimpoi2015deep},
DEP~\cite{xue2018deep}, MAP-Net~\cite{zhai2020dsrnet,zhai2023multiplicity},
DeepTEN~\cite{zhang2017deepten}, and recent
extensions~\cite{evani2025chebyshev,sikdar2025interweaving} routinely
exceed $75\%$ on DTD, $90\%$ on CUB-200, and $97\%$ on Flowers-102. The gap
is attributable not to architecture but to protocol: in those pipelines,
the bulk of discriminative information is already present in the
ImageNet-pretrained representation. Our own results make this contribution
structure visible: ConvNeXt-Tiny trained from scratch on DTD reaches only
$11.1\%$ (Table~\ref{tab:main}, Group~2), although the same architecture
exceeds $75\%$ on DTD when initialized from ImageNet weights. The
${\sim}65$-point difference is therefore overwhelmingly due to large-scale
pretraining.

The from-scratch protocol used here separates two ingredients that
pretrained protocols entangle: the inductive bias of the architecture, and
its ability to learn texture statistics directly from limited
target-domain data. An operator that helps in this regime is encoding a
real bias rather than profiting from the transfer of features learned at
ImageNet scale. The present study makes architectural claims---that
explicit second-order interactions help over implicit learning, that
cross-position displacement adds interpretable directional structure on
top of same-position products, and that AIS-style reweighting is
well-suited to interaction channels---each of which is a claim about what
the network can express from gradient signal on the target dataset alone.
The from-scratch protocol is the cleanest way to test such claims.

The contribution this paper makes is therefore architectural in nature.
Under a protocol that strips away transfer learning, an explicit,
directionally-displaced second-order operator with content-adaptive
selection is a parameter-efficient design that consistently outperforms
both modern attention-based competitors at matched scale and substantially
larger backbones. Combining the operator with ImageNet-pretrained
representations, where the architectural advantage of explicit
cross-position interaction would compound with the appearance prior of a
pretrained backbone, is a complementary direction that we discuss briefly
in Section~\ref{sec:limitations}.

\subsection{Main Results}
\label{sec:main_results}

\begin{figure}[t]
\centering
\includegraphics[width=\columnwidth]{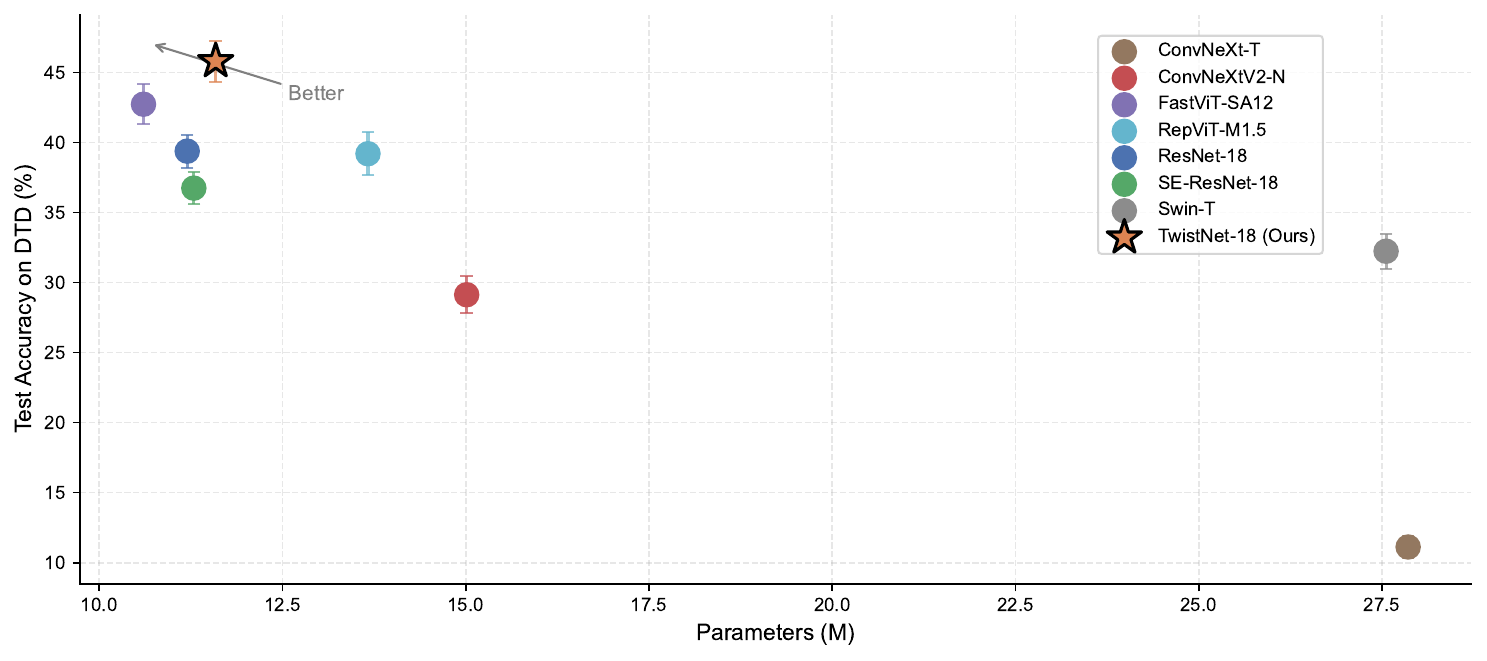}
\caption{\textbf{Accuracy versus parameters on DTD.} \TwistNet-18 attains
the highest accuracy among all models. Group-2 baselines (${\sim}28$M)
degrade sharply without pretraining, illustrating that targeted inductive
bias outweighs raw capacity in data-limited regimes.}
\label{fig:params_acc}
\end{figure}

Main results across the four benchmarks are summarized in
Table~\ref{tab:main} and Fig.~\ref{fig:params_acc}.

\begin{table*}[t]
\centering
\caption{\textbf{Main results.} All models trained from scratch without
ImageNet pretraining under an identical pipeline. Mean accuracy (\%) $\pm$
one standard deviation across folds and three random seeds. Best in
\textbf{bold}, second-best \underline{underlined}. $^\dagger$ Group-2 models
have ${\sim}28$M parameters ($2.4\times$ the parameter-matched group);
their performance under the from-scratch protocol is itself a finding
rather than a deficient training of the baseline.}
\label{tab:main}
\setlength{\tabcolsep}{4pt}
\small
\begin{tabular*}{\textwidth}{@{}l c c c c c c@{}}
\toprule
\textbf{Model} & \textbf{Params} & \textbf{FLOPs} & \textbf{DTD} & \textbf{FMD} & \textbf{CUB-200} & \textbf{Flowers-102} \\
\midrule
\multicolumn{7}{@{}l}{\textit{Group 1: Parameter-matched (10--16M)}} \\
\midrule
ResNet-18~\cite{he2016deep} & 11.20M & 1.82G & 39.4$\pm$1.2 & 42.6$\pm$3.1 & 54.6$\pm$0.5 & 43.6$\pm$0.5 \\
SE-ResNet-18~\cite{hu2018squeeze} & 11.29M & 1.82G & 36.7$\pm$1.2 & 40.8$\pm$2.8 & 52.0$\pm$0.8 & 40.5$\pm$0.7 \\
ConvNeXtV2-Nano~\cite{woo2023convnextv2} & 15.01M & 2.45G & 29.1$\pm$1.3 & 29.7$\pm$2.5 & 31.7$\pm$4.0 & 46.1$\pm$0.6 \\
FastViT-SA12~\cite{vasu2023fastvit} & 10.60M & 1.50G & \underline{42.7$\pm$1.4} & \textbf{45.0$\pm$3.6} & 49.9$\pm$0.6 & \textbf{59.9$\pm$0.6} \\
RepViT-M1.5~\cite{wang2024repvit} & 13.67M & 2.31G & 39.2$\pm$1.5 & 36.6$\pm$2.2 & \underline{59.7$\pm$0.6} & 51.6$\pm$0.7 \\
\rowcolor{gray!15}
\textbf{TwistNet-18 (ours)} & 11.59M & 1.85G & \textbf{45.8$\pm$1.4} & \underline{43.5$\pm$3.8} & \textbf{61.8$\pm$0.5} & \underline{58.5$\pm$0.7} \\
\midrule
\multicolumn{7}{@{}l}{\textit{Group 2: Larger baselines (${\sim}28$M)}} \\
\midrule
ConvNeXt-Tiny$^\dagger$~\cite{liu2022convnet} & 27.86M & 4.47G & 11.1$\pm$0.8 & 24.3$\pm$2.7 & 3.2$\pm$1.4 & 7.5$\pm$0.3 \\
Swin-Tiny$^\dagger$~\cite{liu2021swin} & 27.56M & 4.51G & 32.2$\pm$1.2 & 35.9$\pm$3.2 & 33.0$\pm$1.0 & 48.8$\pm$0.3 \\
\bottomrule
\end{tabular*}
\end{table*}

Among parameter-matched architectures, \TwistNet-18 achieves the highest
accuracy on DTD and CUB-200 and the second-highest on FMD and Flowers-102.
On DTD it improves over ResNet-18 by 6.4 points and over SE-ResNet-18 by
9.1 points. The latter comparison is informative because SE-ResNet-18
places the same SE-style attention used inside MH-STCI on top of the same
backbone, but applies it to first-order features rather than to the
second-order interaction field; the 9.1-point gap therefore isolates the
effect of acting on co-occurrence channels rather than on raw features.
ConvNeXtV2-Nano underperforms even ResNet-18 across all four datasets
despite having 34\% more parameters, consistent with the broader
observation that ConvNeXt-style designs depend heavily on large-scale
pretraining.

The comparison with FastViT-SA12 reveals a regime split rather than a
uniform ranking. \TwistNet shows its largest advantages on DTD ($+3.1$
points) and CUB-200 ($+11.9$ points), benchmarks in which local periodic
patterns---textile weaves, feather barbs---serve as the primary
discriminative cue. FastViT-SA12 retains a small advantage on FMD
($+1.5$) and Flowers-102 ($+1.4$), benchmarks in which material
reflectance and global color distributions play a larger role. This
pattern is consistent with the design intent of STCI: explicit
cross-position correlations are most useful where local periodic
microstructure carries the class signal, and global token mixing remains
advantageous where appearance statistics dominate.

The Group-2 results provide a stress test of the larger architectures
under the from-scratch protocol. ConvNeXt-Tiny falls to $3.2\%$ on CUB-200
(random chance is $0.5\%$ for 200 classes) and $11.1\%$ on DTD, while
Swin-Tiny reaches $33.0\%$ and $32.2\%$. This is consistent with the
broader finding that high-capacity architectures rely on large-scale
pretraining~\cite{dosovitskiy2021vit,liu2022convnet} to generalize on
small target datasets. \TwistNet-18 outperforms ConvNeXt-Tiny by 34.7
points on DTD and 58.6 points on CUB-200 with $2.4\times$ fewer
parameters, indicating that targeted inductive bias outweighs raw capacity
in the data-limited regime.

The fine-grained recognition result on CUB-200 is particularly notable.
\TwistNet-18 reaches $61.8\%$, exceeding the strongest parameter-matched
baseline (RepViT-M1.5) by 2.1 points and FastViT-SA12 by 11.9.
Bird-species discrimination depends substantially on the spacing and
arrangement of plumage micro-structures, which is precisely the case in
which cross-position correlations align peaks of detectors that fire at
periodically displaced positions. On Flowers-102, where color gradients
and overall shape dominate, \TwistNet trails FastViT-SA12 by 1.4 points,
within one standard deviation of either model.

\subsection{Ablation Study}
\label{sec:ablation}

Table~\ref{tab:ablation} ablates the three structural ingredients of
\TwistNet on DTD; each row removes one component while keeping the others
intact, isolating its marginal contribution.

\begin{table}[t]
\centering
\caption{Ablation on DTD. Mean accuracy (\%) $\pm$ one standard deviation
over 10 folds and 3 seeds. Each row removes a single component while
keeping the others intact.}
\label{tab:ablation}
\setlength{\tabcolsep}{12pt}
\renewcommand{\arraystretch}{1.15}
\begin{tabular}{lcc}
\toprule
Variant & Params & Accuracy \\
\midrule
\TwistNet-18 (full)            & 11.59M & \textbf{45.8 $\pm$ 1.4} \\
\midrule
~~w/o Spiral Twist             & 11.59M & 45.6 $\pm$ 1.5 \\
~~w/o AIS                      & 11.53M & 44.1 $\pm$ 1.8 \\
~~First-order only             & 11.20M & 39.4 $\pm$ 1.2 \\
\bottomrule
\end{tabular}
\end{table}

Each row of the table isolates one design choice. The
\emph{first-order only} variant removes both the pairwise products and
the channel-attention reweighting, reducing the block to a standard
ResNet BasicBlock. The \emph{w/o AIS} variant retains the four
directional STCI heads but removes the channel-attention reweighting in
Eq.~\ref{eq:ais}. The \emph{w/o Spiral Twist} variant retains the entire
MH-STCI pipeline---channel reduction, four heads, $\ell_2$-normalization,
pairwise products, AIS, and the gated residual---but replaces the
directional depthwise convolution with the identity, so each head reduces
to a same-position pairwise-product field
$\bar{z}_i(x,y) \cdot \bar{z}_j(x,y)$. This last variant is therefore
informative beyond a simple ablation: it is essentially a local form of
bilinear pooling~\cite{lin2015bilinear,gao2016compact}, the standard
second-order operator, equipped with the same multi-head and AIS
machinery as the full model. It serves as a strong second-order baseline
and isolates the marginal contribution of our novel architectural
component, namely the directional cross-position displacement.

Read in this light, the table tells a coherent story. Replacing the
ResNet block with the local-bilinear-style variant ($w/o$ Spiral Twist)
yields a $39.4 \to 45.6$ improvement of $6.2$ points, which matches the
expected magnitude of the second-order benefit reported across the
bilinear-pooling
literature~\cite{lin2015bilinear,kong2017lowrank,li2017second} and
confirms that explicit pairwise products are an effective inductive bias
in the from-scratch regime. Reintroducing the directional spiral twist on
top of this baseline raises accuracy to $45.8$, the marginal gain of the
operator that distinguishes STCI from a conventional same-position
second-order module. Removing AIS from the full model costs $1.7$ points,
quantifying the value of routing different co-occurrence patterns through
different heads. The decomposition therefore identifies the spiral twist
as the architectural ingredient that takes a strong but conventional
second-order baseline and equips it with cross-position structure.

The size of the spiral-twist increment is itself informative about where
the operator is most useful. DTD spans 47 classes ranging from
\texttt{zigzagged} and \texttt{crystalline} to \texttt{stained} and
\texttt{flecked}, with within-class image-level orientation that varies
nearly arbitrarily under random crops and horizontal flips. In this
heterogeneous regime, orientation-specific signal is distributed across
the four head directions and the channel-attention pathway compensates
for any per-image misalignment by upweighting whichever head matches the
input. A same-position second-order field followed by AIS therefore
already captures most of the discriminative co-occurrence signal on DTD,
because the discriminative orientation enters through the structure of
the local pairwise products themselves rather than through their spatial
offset. The displacement contributes the remaining margin and, more
importantly, supplies the directional structure that makes the four heads
acquire distinct, interpretable orientation selectivities
(Section~\ref{sec:analysis} and Fig.~\ref{fig:interaction}). Without it
the heads collapse to a single same-position computation and the link to
classical co-occurrence analysis that motivates the design is lost.

\subsection{Analysis of Learned Directional Selectivity}
\label{sec:analysis}

\begin{figure}[t]
\centering
\includegraphics[width=\columnwidth]{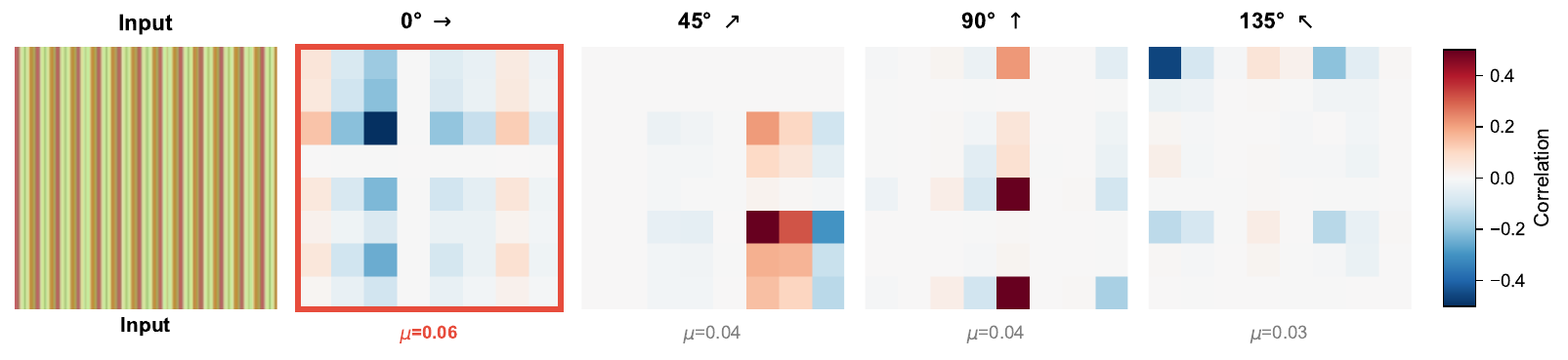}\\[2pt]
\includegraphics[width=\columnwidth]{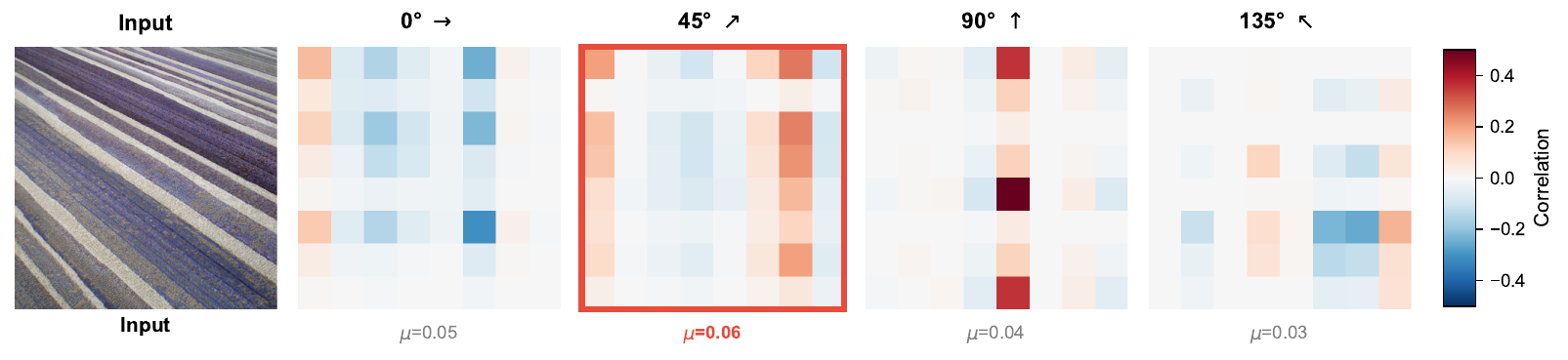}\\[2pt]
\includegraphics[width=\columnwidth]{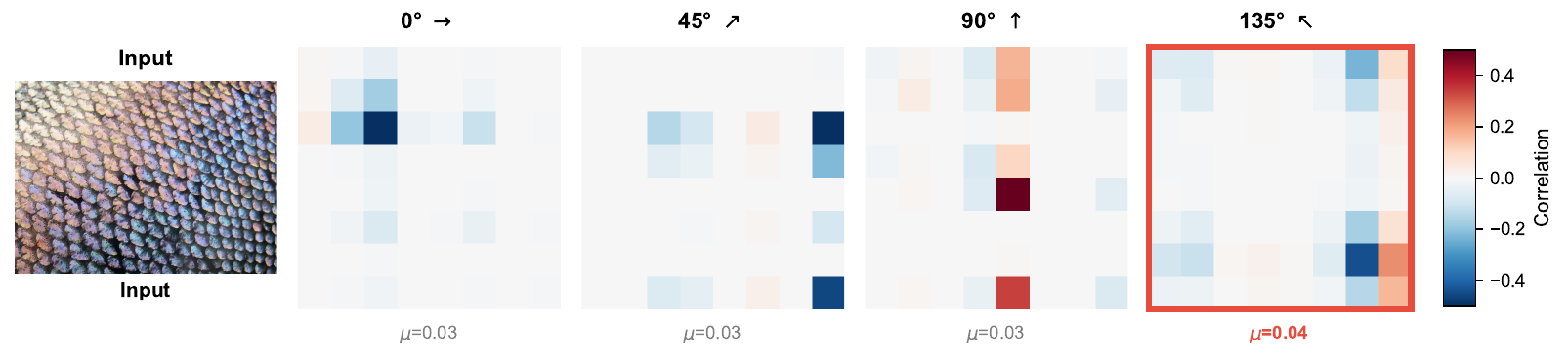}
\caption{\textbf{Learned direction selectivity.} Channel interaction
matrices for three DTD textures. Columns: four directional heads
($0^\circ$, $45^\circ$, $90^\circ$, $135^\circ$). Red borders and bold
$\mu$ values indicate the strongest-responding direction per row. The
values of $\mu$ are channel-pair-mean magnitudes after AIS reweighting,
summarizing the post-attention mass routed through each head.}
\label{fig:interaction}
\end{figure}

Figure~\ref{fig:interaction} visualizes the channel-interaction matrices
of STCI on three DTD samples with distinct dominant orientations. The four
columns correspond to the four directional heads, and the scalar $\mu$
reports the mean magnitude of AIS-reweighted interaction channels for
that head, averaged over all $C_r(C_r{+}1)/2 = 36$ pairs and over all
spatial positions. Two quantities should be distinguished when reading
the figure: $\mu$ measures the post-attention mass routed through the
head, which is the quantity downstream layers consume, while the spatial
heatmap shows where a particular interaction channel within the head
fires. These need not rank directions identically, since a head can have
moderate spatial peaks but a high mean if many channels fire weakly
across the field.

In all three rows, the $\mu$-ranked strongest head corresponds to the
canonical direction aligned with a structural axis of the texture. The
vertical-stripe texture activates the $0^\circ$ head most strongly
($\mu = 0.06$), capturing horizontal co-occurrence perpendicular to the
stripes---the canonical Haralick prescription for vertical-stripe
detection~\cite{haralick1973textural}. The diagonal-stripe texture
activates the $45^\circ$ head ($\mu = 0.06$), aligning with its dominant
orientation. The scaly texture peaks at $135^\circ$ ($\mu = 0.04$),
corresponding to the tilted diagonal arrangement of scales. The spread
of $\mu$ across heads is moderate (the strongest head routes
approximately $1.5\times$ the post-attention mass of the weakest), which
is consistent with AIS amplifying the most informative direction without
collapsing the others; full collapse would make the network fragile to
within-class orientation variation. STCI therefore acquires
orientation-selective co-occurrence detectors without explicit
supervision, in a form that recovers the orientation prescriptions of
classical co-occurrence analysis from gradient signal alone.

\subsection{Regime Analysis}
\label{sec:cases}

The four datasets in Table~\ref{tab:main} span a spectrum of texture
character whose qualitative structure helps explain when \TwistNet's
mechanism is most useful. On CUB-200 and DTD, where dataset-internal
periodicity is strong---plumage micro-structure on the former, oriented
textile patterns on the latter---\TwistNet improves over ResNet-18 by 7.2
and 6.4 points respectively, the largest gains in the study, and matches
or exceeds all parameter-matched baselines. On FMD, where within-class
variability is dominated by lighting, viewing angle, and substrate finish
rather than spatial periodicity, \TwistNet remains competitive with
FastViT-SA12 (43.5 vs.\ 45.0) but does not surpass it; the gap is within
one standard deviation of either model. On Flowers-102, where color
gradients and shape dominate, FastViT-SA12 leads by 1.4 points, again
within one standard deviation, while \TwistNet still outperforms ResNet-18
by 14.9 points, indicating that the gated injection correctly down-weights
the second-order branch when its statistics are not the discriminative
signal. The pattern is consistent with the design intent of STCI: it is
most useful where local periodic structure carries class identity and
gracefully neutral where it does not.

A qualitative inspection of single-fold confusion matrices on DTD confirms
the same pattern at the class level. The classes for which \TwistNet
improves most over ResNet-18 cluster around \texttt{striped},
\texttt{lined}, \texttt{zigzagged}, \texttt{grooved}, \texttt{interlaced},
and \texttt{woven}, all of which are defined by aligned displaced
co-occurrence; gains are smaller on \texttt{stained}, \texttt{cobwebbed},
\texttt{frilly}, and \texttt{fibrous}, which are characterized by
reflectance or shape rather than periodic structure. The class-level
pattern thus mirrors the dataset-level pattern of
Table~\ref{tab:main}: the operator delivers its largest gains where the
class definition rests on aligned displaced co-occurrence, and remains
neutral where it does not.

\section{Discussion and Future Directions}
\label{sec:limitations}

The findings of the present study suggest several natural extensions of
the operator. Combining MH-STCI with ImageNet-pretrained backbones is the
most direct one: the from-scratch evaluation isolates architectural
inductive bias, and integrating that bias with the appearance prior of a
pretrained representation is the cleanest way to position the operator
against the broader texture-recognition literature. The gated-injection
design was deliberately built to be backbone-agnostic, so transfer to
deeper hosts---ResNet-34/50, ConvNeXt-Femto, EfficientNet-B0---should
follow with minimal modification. A second direction is multi-scale
generalization: the present operator uses a fixed displacement, and
pyramidal or learnable variants such as deformable depthwise twists offer
a principled way to capture coarser periodic structure relevant to
textures whose dominant period exceeds a single feature-map pixel.
Finally, because STCI is a drop-in feature-map transform, its integration
into pyramid networks for dense-prediction texture analysis---material
segmentation, surface inspection, remote-sensing land-cover
classification---is a promising application of the same operator to a
different output structure.

\section{Conclusion}
\label{sec:concl}

We presented \TwistNet, a lightweight architecture that incorporates
explicit second-order channel interactions into convolutional networks
through Spiral-Twisted Channel Interaction modules. By computing pairwise
feature correlations along multiple directions and reweighting them
adaptively, \TwistNet captures the co-occurrence statistics that classical
texture theory identifies as the defining signal of structured patterns,
in a form that is end-to-end differentiable and integrable into a standard
ResNet block at negligible cost. Under a controlled, from-scratch
evaluation that isolates architectural contribution from transfer
learning, \TwistNet-18 consistently outperforms parameter-matched
baselines and substantially larger ConvNeXt and Swin Transformer
backbones across four texture and fine-grained recognition benchmarks,
while adding only $3.5\%$ parameters and $2\%$ FLOPs over ResNet-18.
A component-level decomposition shows that the bulk of the gain arises
from explicit local second-order modeling with content-adaptive
selection, and the directional spiral twist supplies the structure that
makes the four heads acquire interpretable, orientation-selective
behavior aligned with classical co-occurrence analysis. Together, these
results indicate that an explicit, spatially-aware second-order operator
is a competitive and theoretically grounded inductive bias for texture
representation, and they invite further study in combination with
pretrained backbones, deeper hosts, and multi-scale extensions of the
operator.

\printcredits

\section*{Statements and Declarations}

\subsection*{Competing interests}
The authors declare that there are no competing interests.

\subsection*{Data availability}
The open code will be available at \url{https://github.com/junbolian/TwistNet-2D}.

\subsection*{Acknowledgments}
This research is financially supported by the National Natural Science
Foundation of China (Grant No.\ U25A20450, 62571374).

\bibliographystyle{model1-num-names}
\bibliography{paper}

\end{document}